\documentclass{article}
\usepackage{spconf,amsmath,graphicx}
\usepackage{spconf,amsmath,graphicx,colortbl}
\usepackage{xcolor}
\usepackage{amssymb}
\usepackage{hyperref}
\hypersetup{hidelinks,
	colorlinks=true,
	allcolors=black,
	pdfstartview=Fit,
	breaklinks=true}

\title{Dual-Path Coupled Image Deraining Network via Spatial-Frequency Interaction}
\name{Yuhong He$^{1}$, Aiwen Jiang†$^{2}$, Lingfang Jiang$^{2}$, Zhifeng Wang$^{3}$, Lu Wang$^{1}$ 
\thanks{
   }
  }

\begin{document}
%
\maketitle
\begin{abstract}
Transformers have recently emerged as a significant force in the field of image deraining. Existing image deraining methods utilize extensive research on self-attention. Though showcasing impressive results, they tend to neglect critical frequency information, as self-attention is generally less adept at capturing high-frequency details. To overcome this shortcoming, we have developed an innovative Dual-Path Coupled Deraining Network (DPCNet) that integrates information from both spatial and frequency domains through Spatial Feature Extraction Block (SFEBlock) and Frequency Feature Extraction Block (FFEBlock). We have further introduced an effective Adaptive Fusion Module (AFM) for the dual-path feature aggregation. Extensive experiments on six public deraining benchmarks and downstream vision tasks have demonstrated that our proposed method not only outperforms the existing state-of-the-art deraining method but also achieves visually pleasuring results with excellent robustness on downstream vision tasks. The source code is available at \href{https://github.com/Madeline-hyh/DPCNet} {https://github.com/Madeline-hyh/DPCNet}.
 

\end{abstract}
\section{Introduction}
\label{sec:intro}

Single image deraining is a challenging task that aims to restore a high-quality image by removing rain degradation from its corresponding degraded input. Rain degradations severely impact the quality of images captured outdoors, which in turn limits the performance of subsequent high-level tasks such as object detection and semantic segmentation \cite{detection}.

Traditional methods rely on diverse image priors through exploring the physical properties of rain streaks \cite{Gauss}. However, these methods have difficulty dealing with complex rainy images in real-world scenarios. Convolutional neural networks (CNNs) have emerged as a preferable choice compared to conventional approaches in recent years due to their local connectivity and translation equivariance \cite{peng, peng2021ensemble, MPRNet}. However, spatial convolution suffers from a limited receptive field, which hinders the establishment of global dependencies and restricts the model's performance. To overcome this limitation, some researchers have attempted to apply vision transformers (ViT) to low-level tasks including image deraining\cite{icip2023_vit_fre, MFDNet}.
 
Although remarkable progress has been made in the single image deraining task, two main limitations still exist. (1) Most of methods primarily exploit spatial information and neglect distinguished frequency information. As demonstrated in \cite{less_effective}, self-attention is less effective in learning high-frequency information. Recently, some methods have considered introducing frequency domain operations for low-level tasks \cite{icip2023_vit_fre, wang2023fourllie}. The research trend gives us an important cue that the frequency domain offers abundant information that should not be overlooked in image-deraining tasks. (2) Most of methods utilize self-attention only along a single dimension (spatial or channel), which limits their effective receptive field to capture more comprehensive interactions from both spatial and channel dimensions.
\begin{figure}[tb]
\small
\centering
\begin{minipage}[b]{1.0\linewidth}
  \centering
\includegraphics[width=1.0\linewidth]
{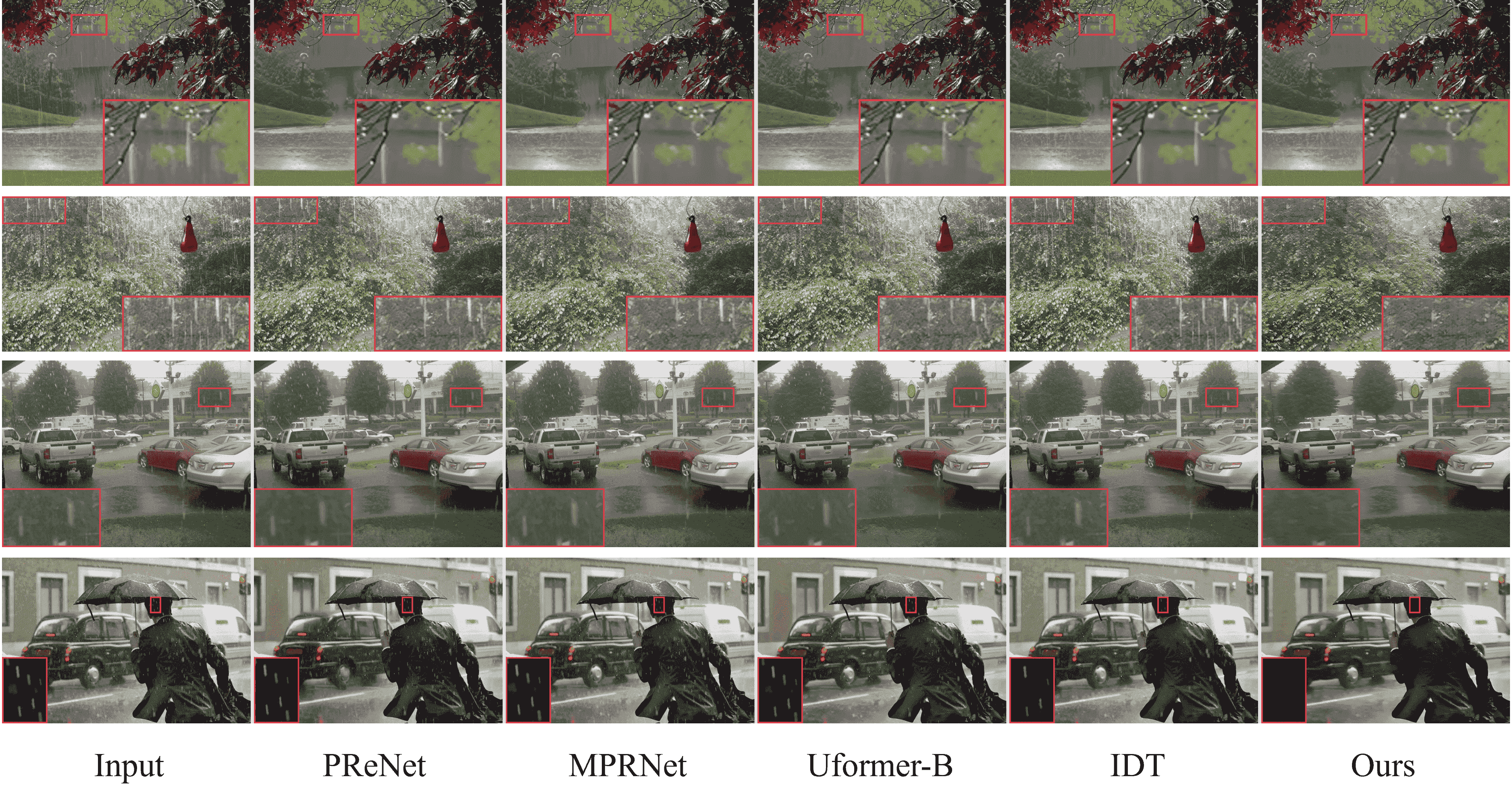}
\end{minipage}
\caption{Visual comparison on the real datasets, including Real15 \cite{JORDER} and Real300 \cite{Real300}. Our method reconstructs credible textures with visually pleasing results. }
\label{Real15}
\end{figure}

To deal with the above-mentioned shortcomings, in this paper, a dual-path coupled image-deraining network (DPCNet) is proposed to enhance the use of spatial-frequency information for the single image deraining task. Specifically, DPCNet is constructed in a multi-scale encoder-decoder structure. It contains several dual-domain blocks (DDBlock) to maximize information interaction between two domains. Each DDBlock consists of a Spatial Feature Extraction Block (SFEBlock), a Frequency Feature Extraction Block (FFEBlock), and an Adaptive Fusion Module (AFM). The SFEBlock is proposed to model pixel-level features from both spatial and channel dimensions simultaneously, mining the potential correlations across the two dimensions. The FFEBlock is based on the Fast Fourier Transform (FFT) to recover high-frequency texture. The AFM is proposed to aggregate the features from two paths adaptively. The above-mentioned structure empowers DPCNet with the capability to capture diverse and intricate feature representations and facilitate rain removal and detailed information restoration.

The main contributions of our work can be summarized as follows:
\begin{itemize}
    \item We propose an effective dual-path coupled image deraining network (DPCNet) via spatial-frequency interaction. The DPCNet facilitates the distinction of rain perturbations and the recovery of backgrounds from both spatial and frequency domains. Extensive experiments demonstrate that our method outperforms the previous state-of-the-art approaches on several public synthetic and real-world datasets.
    \item We propose a Spatial-Channel Transformer Block (SCTB) to model pixel-level features from both spatial and channel dimensions simultaneously, mining the potential correlations across dual dimensions. The SCTB can offer higher receptive field information.
    \item  We extensively evaluate DPCNet across various rain scenes including rain streaks/raindrops and downstream object detection tasks. The extensive experiments have validated the robustness and practicability of the proposed DPCNet.
\end{itemize}

\section{Related work}
\label{sec:format}
Existing single image deraining methods can be divided into two categories: traditional and deep learning-based methods \cite{survey}. Traditional methods mainly introduce various priors to provide additional constraints, such as Gaussian mixture model-based prior, sparsity-based prior, and high-frequency prior for separating rain streaks from clean images. However, these methods can only deal with specific rainy situations, resulting in poor generalization ability in diverse rainy images from different scenes. Recently, numerous deep-learning methods have been proposed and achieved excellent performance \cite{URML, RESCAN, PreNet}. For example, Zhang \textit{et al.} \cite{DIDMDN} proposed a density-aware multi-stream dense network to better characterize rain streaks with different scales and shapes. Jiang \textit{et al.} \cite{MSPFN} proposed to explore the multi-scale collaborative representation of rain streaks. In \cite{MPRNet}, a multi-stage architecture is proposed to learn the restoration function of degraded inputs. 

Recent studies show that transformer-based models have been studied for low-level vision tasks extensively. Chen \textit{et al.} \cite{IPT} propose an encoder-decoder architecture IPT to recover fine-grained predictions. However, these approaches are based on global attention, so they have high computational complexity. Recently, the Swin Transformer leverages local attention with a shifting approach for patch connection and achieves a competitive performance with low complexity. Liang \textit{et al.} propose a strong baseline model SwinIR \cite{swinir} for image restoration based on the Swin Transformer. Wang \textit{et al.} \cite{uformer} introduce a high-performing transformer-based network called Uformer for image restoration. Lee \textit{et al.} \cite{KiT} present an attention mechanism for image restoration, known as k-NN Image Transformer (KiT). Xiao \textit{et al.} \cite{IDT} design an effective transformer-based architecture that can capture long-range and complicated rainy information. Zamir \textit{et al.} \cite{restormer} propose an effective network for image restoration. 

Albeit these methods have shown promising performance in learning the degradation representation from rainy images, they still suffer from limited performance since they few focus on the frequency domain and have restricted representational capacity, which leads to over- and under-enhancement problems. Therefore, a powerful and robust architecture that takes into account both the spatial and frequency domains is of great importance for effective image deraining.

\begin{figure*}[tb]
\small
\centering
\begin{minipage}[b]{1.0\linewidth}
  \centering
\includegraphics[width=1.0\linewidth]{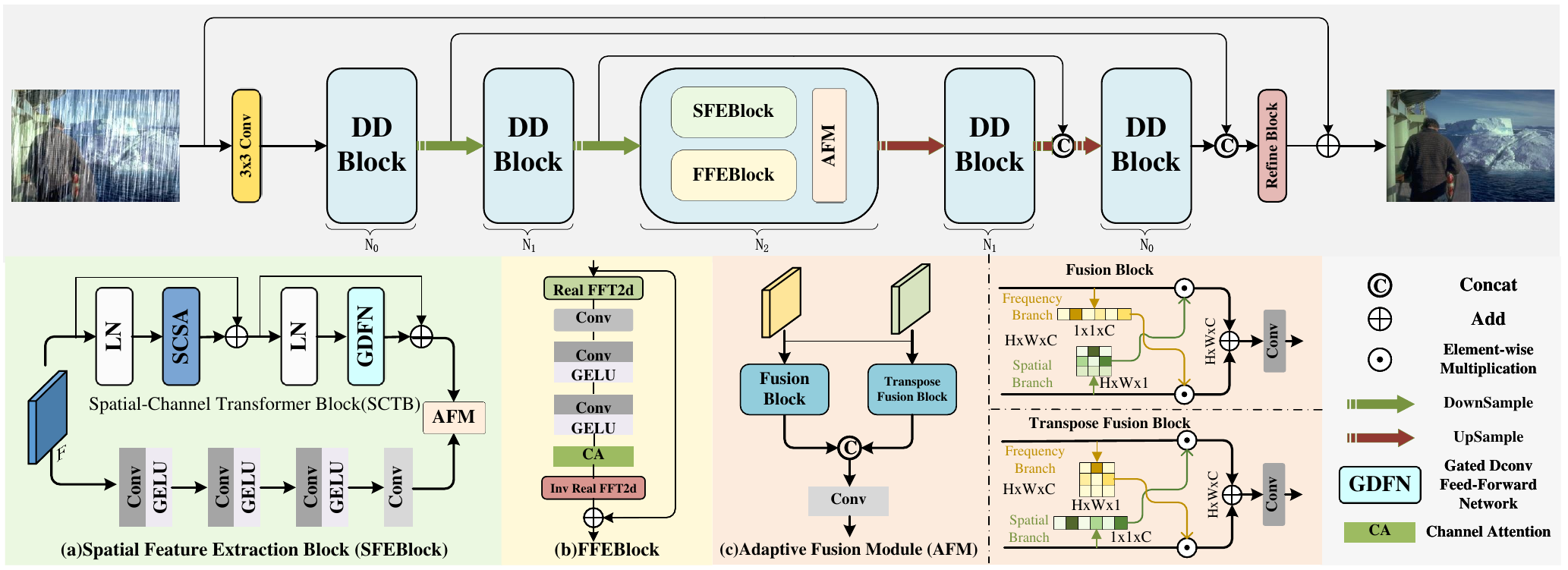}
\end{minipage}
\caption{Our proposed DPCNet consists of several Dual-Domain Blocks (DDBlock). The DDBlock is proposed to extract features from both spatial and frequency domains. The Adaptive Fusion Module (AFM) is proposed to interact dimensional information between two branches by reweighting features from the spatial and channel dimensions alternately.}
\label{network}
\end{figure*}

\begin{figure}[tb]
\small
\centering
\begin{minipage}[b]{1.0\linewidth}
  \centering
\includegraphics[width=1.0\linewidth]
{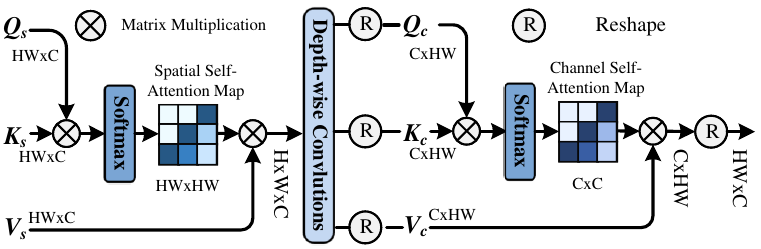}
\end{minipage}
\caption{Illustrations of our proposed Spatial-Channel Self-Attention (SCSA), which contains spatial-based self-attention and channel-wise self-attention. }
\label{transformer}
\end{figure}

\section{PROPOSED METHOD}
\label{sec:method}
The architecture of DPCNet is shown in Fig. \ref{network}. It contains several dual-domain blocks (DDBlock) within the encoder-decoder framework. The proposed DDBlock comprises three modules: Spatial Feature Extraction Block (SFEBlock), Frequency Feature Extraction Block (FFEBlock), and Adaptive Fusion Module (AFM). 

\subsection{Spatial Feature Extraction Block}
\label{ssec:subhead}
As shown in Fig. \ref{network} (a), the SFEBlock is a well-designed asymmetric hybrid dual-path structure for global clean scene reconstruction and local detail feature enhancement. It contains a specifically designed Spatial-Channel Transformer Block (SCTB) for aggregating spatial and channel features to obtain powerful representation ability. We also introduce convolutions to enhance locality while capturing long-range dependencies. Since simply concatenating or adding the dual-branch features cannot effectively couple global and local features, the two-branch features are aggregated using the Adaptive Fusion Module (AFM).


\noindent{\bf Spatial-Channel Transformer Block.} As depicted in Fig. \ref{transformer}, unlike existing self-attention paradigms that only indulge in spatial \cite{uformer, IDT} or channel \cite{restormer, MFDNet} dimensional processing, the SCTB are based on spatial-window\cite{uformer, IDT} and channel-wise self-attention simultaneously. Through aggregating two types of self-attention, SCTB can realize information transmission between two dimensions in a compact manner. 

Specifically, in case of spatial-window self-attention, as displayed in Fig. \ref{transformer}, we first generate query $Q^s$, key $K^s$, and value $V^s$ matrices through linear projection, and partition $Q^s$, $K^s$, and $V^s$ into non-overlapping local windows. The process is defined in Equation~\ref{eq:kQV}. 
\begin{equation}
\label{eq:kQV}
Q^s=X W_{Q}, K^s=X W_{K}, V^s=X W_{V}
\end{equation}
where, $W_{Q}$, $W_{K}$, $W_{V}$ $\in$  $\mathcal{R}^{C \times C}$ are linear projection operations. 
Then, we get the flattened features from each window and split them into $h$ heads. We calculate the production of query and key to get the spatial attention map of size $\mathcal{R}^{HW \times HW}$. The output $F_{i}^{s}$ for the $i$-th head is defined as in Equation~\ref{multi-head}.
\begin{equation}
\begin{array}{l}
F_{i}^{s}=\operatorname{softmax}\left(Q_{i}^{s}\left(K_{i}^{s}\right)^{T} / \sqrt{d}\right) \cdot V_{i}^{s}  \\
F^{s}=\operatorname{concat}\left(F_{1}^{s},...,F_{h}^{s}\right)
\end{array}
\label{multi-head}
\end{equation}
where, $d= C/h$ means the dimension of each head $h_{i}$. Finally, we obtain the intermediate aggregated results $F^{s}$ by concatenating the outputs for all heads.

In case of channel-wise self-attention, it applies multi-head SA across channels. Specifically, we first reshape the input query, key, and value matrix $Q^c$, $K^c$, and $V^c$ $\in$  $\mathcal{R}^{C \times HW}$ to obtain the attention map of size $\mathcal{R}^{C \times C}$, which can model channel-wise relationships and encoder the global context implicitly. Then, we divide channels into heads and apply attention per head separately. The process of the final output $F^{c}$ is the same as Equation \ref{multi-head}. All of the Feed-Forward Network (FFN) in the SCTB is Gated Dconv Feed-Forward Network (GDFN) \cite{restormer}, as shown in Fig. \ref{network} (a).

\begin{table*}[tb]
\centering
\caption{Quantitative PSNR($\uparrow$) and SSIM($\uparrow$) comparisons with existing state-of-the-art deraining methods. Average means the average performance of the five benchmark datasets. The \textbf{bold} and \underline{underline} represent the best and second-best performance. }
\label{quantitative} 
\resizebox{1.0\linewidth}{!}{
\begin{tabular}{ccccccccccc|
>{\columncolor[HTML]{EFEFEF}}c 
>{\columncolor[HTML]{EFEFEF}}c }
\hline
                & \multicolumn{2}{c}{Test100} & \multicolumn{2}{c}{Rain100H} & \multicolumn{2}{c}{Rain100L} & \multicolumn{2}{c}{Test1200} & \multicolumn{2}{c|}{Test2800} & \multicolumn{2}{c}{\cellcolor[HTML]{EFEFEF}Average} \\
Metrics         & PSNR         & SSIM         & PSNR          & SSIM         & PSNR          & SSIM         & PSNR          & SSIM         & PSNR          & SSIM          & PSNR                     & SSIM                     \\ \hline
DerainNet \cite{DerainNet}       & 22.77        & 0.810        & 14.92         & 0.592        & 27.03         & 0.884        & 23.38         & 0.835        & 24.31         & 0.861         & 22.48                    & 0.796                    \\
DIDMDN \cite{DIDMDN}          & 22.56        & 0.818        & 17.35         & 0.524        & 25.23         & 0.741        & 29.95         & 0.901        & 28.13         & 0.867         & 24.64                    & 0.770                    \\
UMRL \cite{URML}            & 24.41        & 0.829        & 26.01         & 0.832        & 29.18         & 0.923        & 30.55         & 0.910        & 29.97         & 0.905         & 28.02                    & 0.880                    \\
RESCAN \cite{RESCAN}          & 25.00        & 0.835        & 26.36         & 0.786        & 29.80         & 0.881        & 30.51         & 0.882        & 31.29         & 0.904         & 28.59                    & 0.858                    \\
SPANet \cite{SPANet}          & 23.17        & 0.833        & 26.54         & 0.843        & 32.20         & 0.951        & 31.36         & 0.912        & 30.05         & 0.922         & 28.66                    & 0.892                    \\
PReNet \cite{PreNet}          & 24.81        & 0.851        & 26.77         & 0.858        & 32.44         & 0.950        & 31.36         & 0.911        & 31.75         & 0.916         & 29.43                    & 0.897                    \\
MSPFN \cite{MSPFN}           & 27.50        & 0.876        & 28.66         & 0.860        & 32.40         & 0.933        & 32.39         & 0.916        & 32.82         & 0.930         & 30.75                    & 0.903                    \\
PCNet \cite{PCNet}           & 28.94        & 0.886        & 28.38         & 0.870        & 34.19         & 0.953        & 31.82         & 0.907        & 32.81         & 0.931         & 31.23                    & 0.909                    \\
MPRNet \cite{MPRNet}          & 30.27       & 0.897        & 30.41         & 0.890        & 36.40         & 0.965        & 32.91         & 0.916        & 33.64         & 0.938        & 32.73                   & 0.921                    \\
DGUNet \cite{DGUNet}       & 30.32        & 0.899        & \underline{30.66}         & 0.891        & 37.42         & 0.969        & \underline{33.23}        & 0.920        & 33.68         & 0.938         & 33.06                    & 0.923  \\
KiT \cite{KiT}       & 30.26        & 0.904        & 30.47         & 0.897        & 36.65         & 0.969        & 32.81         & 0.918        & \underline{33.85}         & \underline{0.941}         & 32.81                    & 0.929  \\
Uformer-B \cite{uformer}       & 28.71        & 0.896        & 27.54         & 0.871        & 35.91         & 0.964        & 32.34         & 0.913        & 30.88         & 0.928         & 31.08                    & 0.914                    \\
IDT \cite{IDT}             & 29.69        & 0.905        & 29.95         & 0.898        & 37.01        & 0.971       & 31.38         & 0.908        & 33.38         & 0.937         & 32.28                    & 0.924                    \\
HPCNet \cite{HPCNet}         & 30.87        & 0.914        & 30.12         & 0.893        & 37.01         & 0.972        & 32.63         & 0.915        & -             & -             & 32.66                    & 0.923                    \\
DAWN \cite{jiang2023dawn}         & 29.86        & 0.902        & 29.89         & 0.889        & 35.97         & 0.963        & 32.76         & 0.919       & -             & -             & 32.12                    & 0.918                    \\
MFDNet \cite{MFDNet}         & \textbf{30.78}        & \textbf{0.914}        & 30.48         & \underline{0.899}        & \underline{37.61}         & \underline{0.973}       & 33.01         & \underline{0.925}       & 33.55             & 0.939            & \underline{33.08}                   & \underline{0.930}                    \\
DPCNet(Ours)   & \underline{30.59}       & \underline{0.914}        & \textbf{30.73}         & \textbf{0.899}        & \textbf{37.96}        & \textbf{0.974}       & \textbf{33.23}         & \textbf{0.928}        & \textbf{33.87}         & \textbf{0.941}         & \textbf{33.28}                   & \textbf{0.931}                    \\ \hline
\end{tabular}
}
\end{table*}

\begin{table*}[tb]
\caption{Quantitative NIQE($\downarrow$)/BRISQUE($\downarrow$) performance comparisons on the real-world datasets.}
\centering
\label{NIQE} 
\setlength{\tabcolsep}{5pt}
\resizebox{1.0\linewidth}{!}{
\begin{tabular}{ccccccccc}
\hline
Datasets    & UMRL \cite{URML}          & PReNet \cite{PreNet}       & MSPFN \cite{MSPFN}         & PCNet \cite{PCNet}        & MPRNet \cite{MPRNet}        & Uformer-B \cite{uformer}    & IDT \cite{IDT}          & DPCNet(Ours) \\ \hline
Real15 \cite{JORDER}      & 16.60/24.09 & 16.04/25.29 & 17.03/23.60 & 16.19/25.61 & 16.48/23.92 & 15.71/18.67 & 15.60/28.35 & \textbf{14.58}/\textbf{18.46} \\
Real300 \cite{Real300}     & 15.78/26.39 & 15.12/23.57 & 15.34/28.27 & 15.47/28.99 & 15.08/28.69 & 14.90/24.35 & 15.45/\textbf{23.48} & \textbf{14.76}/25.22 \\
RID \cite{RID}         & 12.19/40.67   & 12.45/38.89    & 11.74/41.88          & 12.03/40.76    & 11.79/44.04   & 11.54/36.38   & 12.49/\textbf{35.99}   & \textbf{10.96}/37.32             \\
RIS \cite{RID}        & 16.30/47.98   & 16.74/48.83    & 15.73/47.45 & 16.19/49.00    & 16.88/52.11   & 16.00/\textbf{45.06}  & 17.45/49.86   & \textbf{14.54}/47.04          \\     
\hline    
\end{tabular}
}
\end{table*}

\subsection{Frequency Feature Extraction Block}
\label{ssec:subhead}
It is inadequate to estimate the clean image only from a single rainy input in the spatial domain because rain degradations and background are usually highly coupled in the spatial domain. Moreover, the reconstruction process includes low and high-frequency information recovery and SA usually ignores high-frequency information. Thus, we propose the FFEBlock to prioritize the frequency-level information for decoupling derained images. We utilize channel-wise Fast Fourier Transform (FFT) to effectively learn high-frequency information and decouple rain degradation. The process is depicted in Fig. \ref{network} (b). The FFEBlock enjoys benefits from modeling both high-frequency and low-frequency discrepancies between the rainy image and clean image pairs. 


\subsection{Adaptive Fusion Module}
\label{ssec:subhead}
Incorporating different dimensional information within simple concatenation or adding operations is often ineffective. To address this challenge, we propose an AFM to facilitate the interaction of features between two branches. As illustrated in Fig. \ref{network} (c), the AFM reweights features of two branches from the spatial and channel dimensions alternately, as depicted in Equation \ref{AFM}.
\begin{equation}
\begin{array}{l}
F_{1}=\operatorname{conv} ({B}_{1} \odot \mathrm{SA}({B}_{2})+ {B}_{2}\odot \mathrm{CA}({B}_{1}))  \\
F_{2}=\operatorname{conv} ({B}_{1} \odot \mathrm{CA}({B}_{2})+ {B}_{2}\odot \mathrm{SA}({B}_{1}))  \\
F=\operatorname{conv}(\operatorname{concat}\left(F_{1},F_{2}\right))
\end{array}
\label{AFM}
\end{equation}
where, ${B}_{1}$ and ${B}_{2}$ represent the different branches of the network. $\mathrm{SA}$ and $\mathrm{CA}$ represent the spatial attention map and channel attention map generation processes, respectively. $\odot$ represents element-wise dot-production. Subsequently, through the proposed adaptive interaction, the outputs of the two branches can be adjusted to complement each other, thereby achieving improved feature fusion.

\subsection{Loss Function}
\label{ssec:subhead}
The total training loss $\mathcal{L}_{\text {total }}$ can be formulated as following Equation \ref{loss}. As Equation \ref{loss} shows, the total loss consists of L1 loss $\mathcal{L}_{1}$, perceptual loss $\mathcal{L}_{perceptual}$ and FFT loss $\mathcal{L}_{fft}$. The perceptual loss is computed by the first and the third layers of VGG19. 
\begin{equation}
\label{loss}
\begin{aligned}
\mathcal{L}_{total} & =\lambda_{1} \mathcal{L}_{1}\left(\mathcal{B}, \mathcal{B}_{g t}\right)+\lambda_{2} \mathcal{L}_{perceptual}\left(\mathcal{B}, \mathcal{B}_{g t}\right) \\
& +\lambda_{3} \mathcal{L}_{fft}\left(\mathcal{B}, \mathcal{B}_{g t}\right)
\end{aligned}
\end{equation}
where, $\mathcal{B}$ and $\mathcal{B}_{g t}$ denote the predicted de-raining result and the corresponding ground truth. $\lambda_{1}$, $\lambda_{2}$ and $\lambda_{3}$ denote balancing parameters. In our experiment, we set $\lambda_{1}$, $\lambda_{2}$ and $\lambda_{3}$ to 1, 0.2 and 0.05 experimentally. 

\begin{figure*}[tb]
\centering
\includegraphics[width=0.95\textwidth,height=0.45\textwidth]{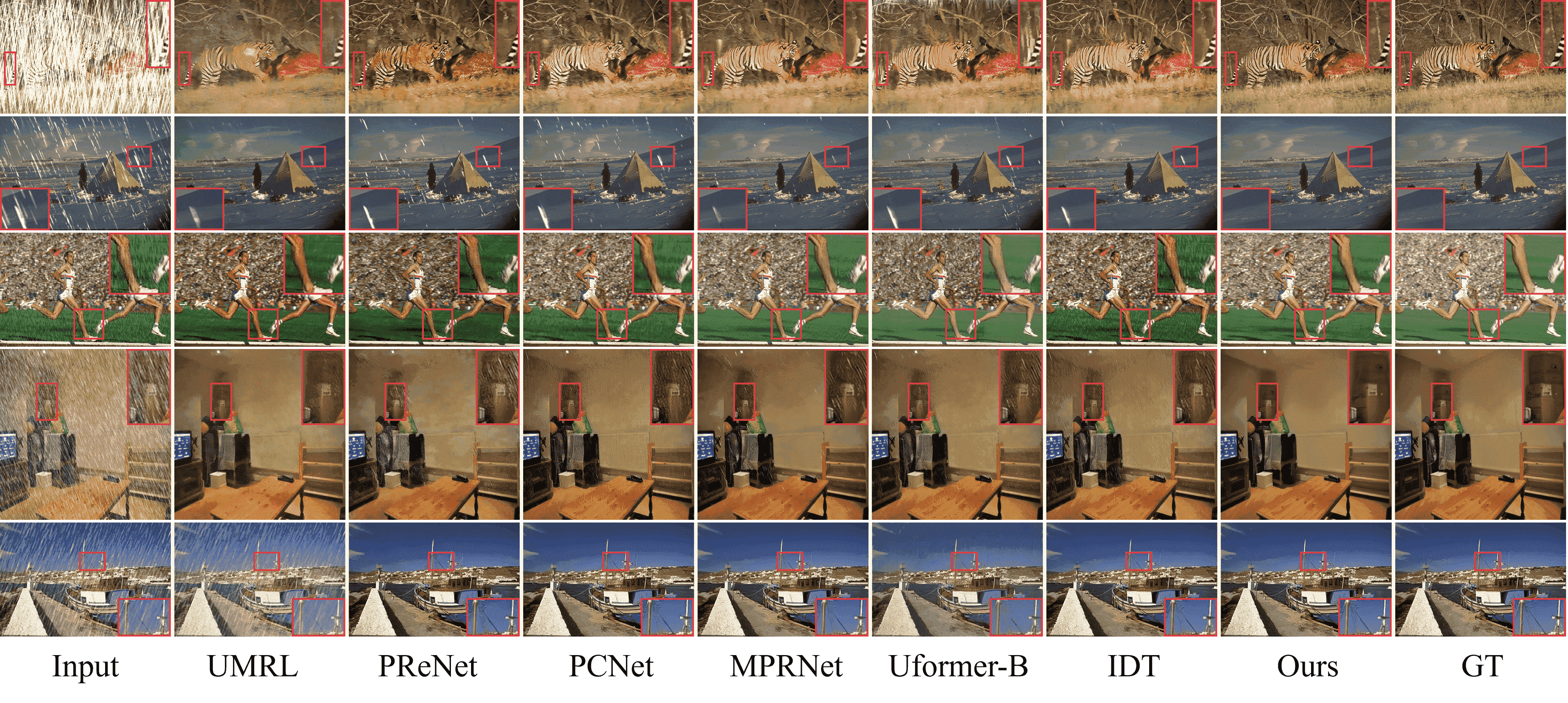}
\centering
\setlength{\abovecaptionskip}{-5pt} 
\caption{Visual comparison on the Rain100H \cite{JORDER}, Rain100L \cite{JORDER}, Test100 \cite{Rain800}, Test1200 \cite{DIDMDN}, and Test2800 \cite{DerainNet} datasets. }
\label{rain13k}
\end{figure*}

\begin{table}[tb]
\small
\caption{Quantitative comparisons with existing state-of-the-art methods on the RainDS \cite{CCN} benchmark, including all types of precipitation (i.e., rain streaks (RS), raindrops (RD), and a combination of both (RDS)). The \textbf{bold} and \underline{underline} represent the best and second-best performance. }
\label{RainDS} 
\resizebox{\columnwidth}{!}{
\begin{tabular}{ccccccc|cc}
\hline
\multicolumn{1}{c}{}                         & \multicolumn{2}{c}{RS}                               & \multicolumn{2}{c}{RD}                               & \multicolumn{2}{c|}{RDS} 
& \multicolumn{2}{c}{Average} \\
{Methods} & \multicolumn{1}{c}{PSNR}  & \multicolumn{1}{c}{SSIM}  & \multicolumn{1}{c}{PSNR}  & \multicolumn{1}{c}{SSIM}  & \multicolumn{1}{c}{PSNR}  & \multicolumn{1}{c|}{SSIM} &
\multicolumn{1}{c}{PSNR}  & \multicolumn{1}{c}{SSIM} 
\\ \hline
DerainNet \cite{DerainNet}                                & 30.41 & 0.869 & 27.92 & 0.885 & 26.85 & 0.796 & 28.39 & 0.850\\
RESCAN \cite{RESCAN}                             & 30.99 & 0.887 & 29.90 & 0.907 & 27.43 & 0.818 & 29.44 & 0.871 \\
PReNet \cite{PreNet}                            & 36.63 & 0.968 & 34.58 & 0.964 & 32.21 & 0.934 & 34.47 & 0.955\\
UMRL \cite{URML}                                & 35.76                     & 0.962                     & 33.59                     & 0.958                     & 31.57                     & 0.929  & 33.64 & 0.950                   \\
JORDER-E \cite{JORDER}                            & 33.65                     & 0.925                     & 33.51                     & 0.944                     & 30.05                     & 0.870      & 32.40 & 0.913               \\
MSPFN \cite{MSPFN}                               & 38.61                     & 0.975                     & 36.93                     & 0.973                     & 34.08                     & 0.947    & 36.54 & 0.965                 \\
CCN \cite{CCN}                                 & 39.17                     & 0.981                     & 37.30                     & 0.976                     & 34.79                     & 0.957    & 37.09 & 0.971                 \\
MPRNet \cite{MPRNet}                              & 40.81                     & 0.981                     & 37.03                     & 0.972                     & 34.99                     & 0.956   & 37.61 & 0.970                  \\
DGUNet \cite{DGUNet}                               &  41.09               & 0.983               &  37.56               & 0.975              & 35.34               & 0.959    & 38.00 & 0.972          \\
Uformer-B \cite{uformer}                              & 40.69                     & 0.972                     & 37.08                     & 0.966                     & 34.99                     & 0.954   & 37.59 & 0.964                  \\
Restormer \cite{restormer}                              & 41.42                     & 0.980                     & 38.78                     & 0.976                     & 36.08                     & \underline{0.961}        & 38.76 & 0.972            \\
IDT \cite{IDT}                              & \underline{41.61}                     & \underline{0.983}                     & \textbf{39.09}                     & \underline{0.980}                     & \underline{36.23}                     & 0.960  & \underline{38.98} & \underline{0.974}                  \\
NAFNet \cite{NAFNet}                               & 40.39                     & 0.972                     & 37.23                     & 0.974                     & 34.99                     & 0.957     & 37.54 & 0.968                \\
DPCNet(ours)                          & \textbf{46.91}            & \textbf{0.994}            & \underline{38.95}            & \textbf{0.984}            & \textbf{37.08}            & \textbf{0.977}        &    
\textbf{40.98}            & \textbf{0.985}  
\\ \hline

\end{tabular}
}
\end{table}

\section{Experiments}
\label{sec:illust}
\subsection{Implementation Details}
\label{ssec:subhead}
The network is implemented in Pytorch, and trained on eight NVIDIA GeForce RTX 3090 GPUs. We use Adam optimizer with an initial learning rate of $3 \times 10^{-4}$ for training. The learning rate of Adam is steadily decreased to $1 \times 10^{-6}$  using the cosine annealing. Following \cite{restormer}, we utilize progressive learning for training and start with a patch size of 128 $\times$ 128 and a batch size of 64.  Our network employs a 3-level encoder-decoder, and the number of Transformer blocks is set to [2,3,4], and self-attention heads are [2,4,8].

\subsection{Datasets and Compared Methods}
\label{ssec:subhead}
{\bf Rain Streak Datasets.} Following \cite{MSPFN, MPRNet}, we conduct experiments on the mainstream de-raining Rain13k dataset. It contains 13,712 images with rain streaks of various scales and directions for training. Test100 \cite{Rain800}, Rain100H \cite{JORDER}, Rain100L \cite{JORDER}, Test1200 \cite{DIDMDN} and Test2800 \cite{DerainNet} are utilized for test. 

\noindent{\bf Raindrops and Rain Streak Datasets.} We evaluate the robustness of the proposed method to cope with raindrop and rain streaks scenarios on the RainDS dataset \cite{CCN}. It contains rain streaks (RS), raindrops (RD), and both (RDS). RainDS has 3600 image pairs, with 3000 images used for training and the remaining 600 images for testing.

\noindent{\bf Real Datasets.} Real datasets are also considered to test the generalization and the effect on downstream object detection tasks, including Real15 \cite{JORDER}, Real300 \cite{Real300}, Rain in Driving (RID), and Rain in Surveillance (RIS) \cite{RID}. RID and RIS have a total of 2495 and 2348 images in real scenes, respectively. 

\noindent{\bf Compared Methods and Evaluation Metrics.} We compare our method with existing several state-of-the-art deraining methods on the mainstream benchmarks. The compared methods include DerainNet \cite{DerainNet}, DIDMDN \cite{DIDMDN}, UMRL \cite{URML}, RESCAN \cite{RESCAN}, SPANet \cite{SPANet}, PReNet \cite{PreNet}, JORDER \cite{JORDER}, MSPFN \cite{MSPFN}, PCNet \cite{PCNet}, CCN \cite{CCN}, MPRNet \cite{MPRNet}, DGUNet \cite{DGUNet}, KiT \cite{KiT}, Uformer-B \cite{uformer}, Restormer \cite{restormer}, IDT \cite{IDT}, NAFNet\cite{NAFNet}, HPCNet \cite{HPCNet}, DAWN \cite{jiang2023dawn}, and MFDNet \cite{MFDNet}. Continuing along the trajectory of previous works \cite{IDT, MPRNet, PreNet}, we use PSNR and SSIM to evaluate the deraining performance of synthetic images on the Y channel of the transformed YCbCr color space, and utilize NIQE and BRISQUE to evaluate the real dataset. For the downstream detection task, we employ commonly used evaluation metrics mAP for overall performance, as well as the F1, Recall, Precision, and AP for specific categories.
%

\begin{table*}[htbp]
    \caption{Object detection results on the RID \cite{RID}.}
	\centering
	
	\label{detection} 
    \resizebox{\linewidth}{!}{
    \begin{tabular}{ccccccccccccc|c}
\hline
& \multicolumn{4}{c}{Car}           & \multicolumn{4}{c}{Person}        & \multicolumn{4}{c|}{Motorbike}  &  \\
                       Methods   & F1   & Recall & Precision & AP    & F1   & Recall & Precision & AP   & F1   & Recall & Precision & AP
                       &  mAP                    \\ \hline
Input                    & 0.71 & 67.11  & 75.31     & 65.51 & 0.52 & 62.4   & 45.18     & 48.8  & 0.22 & 14.12 & 53.60 & 14.48 & 43.09                \\
UMRL \cite{URML}                     & 0.74 & 72.65  & 75.21     & 70.3  & 0.53 & 63.83  & 44.97     & 50.59 & 0.26&\underline{16.74}&54.39&\underline{17.37}&46.09                \\
PReNet  \cite{PreNet}                 & 0.73 & 70.83  & \underline{75.88}    & 69.12 & 0.51 & 55.64  & \textbf{46.75}    & 46.39 & 0.24&15.51&51.39&15.91&40.92                \\
MSPFN  \cite{MSPFN}                & 0.74 & 71.86  & 75.31     & 70.15 & 0.54 & 67.02  & 45.15     & 52.41 & 0.24&15.44&53.60&15.83&46.43                \\
PCNet  \cite{PCNet}                  & 0.70  & 72.35  & 67.34     & 63.02 & 0.49 & 66.78  & 38.83     & 45.68 & 0.24&16.11&48.44&14.24&40.78                \\
MPRNet  \cite{MPRNet}                 & 0.74 & \underline{72.79}  & 74.87     & 70.40 & 0.53 & \underline{67.14}  & 43.64     & 52.22 & 0.23&15.06&53.11&15.66&45.89                \\
Uformer-B  \cite{uformer}              & 0.74      & 72.09       &  75.53         & 70.26      &  0.53    &   64.66     &   44.98        &   50.88    &  0.26&16.55&\underline{56.99}&15.46&45.62                    \\
IDT  \cite{IDT}                    & \underline{0.74} & 72.75  & 75.04     & \underline{70.77} & \underline{0.54} & 66.61  & 45.11     & \underline{52.48} & \underline{0.26}&\textbf{16.84}&54.92&16.64&\underline{46.44}                \\
DPCNet(Ours)                     & \textbf{0.78} & \textbf{76.72}  & \textbf{77.69}    & \textbf{74.32} & \textbf{0.54} & \textbf{68.29}  & \underline{45.25}     & \textbf{55.41} & \textbf{0.26}&16.11&\textbf{63.72}&\textbf{17.98}&\textbf{47.46}               \\ \hline
\end{tabular} 
}
\end{table*}

\begin{table}[]
\caption{Ablation studies on Rain100H. The first raw is to study the necessity of each component of the network. The second raw is to verify the reasonability of the loss function.}
	\centering
	\label{ablation} 
 \setlength{\tabcolsep}{6pt}
    \resizebox{\columnwidth}{!}{
\begin{tabular}{c|ccc|cc}
\hline
  Methods & L1        & FFT       & Perceptual & PSNR & SSIM      \\ \hline
(1)Single spatial domain branch & \checkmark             & \checkmark         & \checkmark         & 30.17      & 0.891          \\
(2)Replace AFM with concatenation & \checkmark              & \checkmark         & \checkmark         & 30.57      & 0.897         \\ 
(3)Only spatial-based SA in SCTB & \checkmark             & \checkmark         & \checkmark         & 30.43      & 0.897          \\ 
(4)Only channel-wise SA in SCTB & \checkmark             & \checkmark         & \checkmark         & 30.22      & 0.896          \\ 
(5)Change the order of the two types of SA & \checkmark              & \checkmark         & \checkmark         & 30.42      & 0.896          \\ 
(6)(Ours) & \checkmark  &    \checkmark      & \checkmark                   & \textbf{30.73}      & \textbf{0.899}          \\ \hline

(7)(ours) & \checkmark  &                      &            & 30.36      &  0.895         \\
(8)(ours) & \checkmark             & \checkmark         &           & 30.61      & 0.897  \\
(9)(ours) & \checkmark             &  & \checkmark          & 30.52     &   0.897       
        \\ \hline
\end{tabular}
}
\end{table}

\subsection{Quantitative and Qualitative Experiment}
\label{ssec:subhead}
{\bf Synthetic Images.} 
The comparison results are shown in Tab. \ref{quantitative}. Our method achieved the best results on the average performance of five test datasets than previous SOTA methods. Compared to the second-best method MFDNet\cite{MFDNet}, our method has 0.2 dB gain on PSNR. The visual comparisons are shown in Fig. \ref{rain13k}. Compared with existing methods, our approach removes rain streaks more completely and restores better texture details of the background, while other approaches retain some obvious rain streaks or lose important details of the background. 

Moreover, we conducted additional experiments to evaluate the robustness of our method on the RainDS \cite{CCN} benchmark. The quantitative comparison results are presented in Tab. \ref{RainDS}. Among all the models, our method achieves remarkable performance improvements, indicating that it can remove raindrops and rain streaks uniformly.

\noindent{\bf Real Images.} We further conduct experiments on four real-world datasets: Real15 \cite{JORDER}, Real300 \cite{Real300}, RID \cite{RID}, and RIS \cite{RID}. Quantitative results of NIQE and BRISQUE are listed in Tab. \ref{NIQE}. Our method achieves the best NIQE performance on all of the four datasets, which demonstrates that our method achieves good generalization in real-world scenarios. Visual comparisons on Real15 and Real300 are illustrated in  Fig.\ref{Real15}.

\subsection{Ablation Studies}
\label{ssec:subhead}
To evaluate the effectiveness of each proposed component, we conduct ablation studies on the Rain100H dataset. As shown in Tab. \ref{ablation}, we can observe that all components are crucial for our architecture. The ablation cases are: (1) Single spatial domain branch: only SFEBlock in the DDBlock. (2) Using concatenation to replace the AFM; (3) Only spatial-based SA in SCTB: without channel-wise SA. (4) Only channel-wise SA in SCTB: without spatial-based SA. (5) Change the order of the two types of SA: channel-wise self-attention followed by spatial-based self-attention. 

In case of (1) and (2), the performance of the proposed method degrades 0.56dB and 0.008 on PSNR and SSIM without the frequency information. It demonstrates that frequency domain information and adaptive fusion are indispensable for single image deraining. The performance degrades 0.16dB on PSNR without the AFM, which demonstrates that adaptive information interaction is critical to removing rainy patterns. (3)-(5) demonstrates that our proposed SCTB is effective in removing rainy patterns. The ablation results of the loss function are also shown in Tab. \ref{ablation}, which demonstrates that our loss function is effective for single image de-raining.

\subsection{Downstream Object Detection Task on Derain Image}
\label{ssec:subhead}
To explore the effect of deraining methods on downstream vision tasks, we directly apply the pre-trained YOLOv7 model to the RID dataset \cite{RID} for object detection. The RID consists of five classes: car, person, motorbike, bus, and bicycle. For our analysis, we focus on one of the three largest classes and calculate its F1, Recall, Precision, and AP results,  as presented in Tab. \ref{detection}. As a result, we compute the mean Average Precision(mAP) results of all classes on the RID, and our method achieves the best performance. The experiment results demonstrate that our method is capable of benefiting the downstream object detection task effectively. 

\section{Conclusion}
\label{conclusion}
In this paper, an effective dual-path coupled image deraining network (DPCNet) has been proposed. In spatial path, spatial and channel self-attention are utilized simultaneously, mining potential interactions between the two directions. In frequency path, the Fast-Fourier Transform (FFT) is utilized to extract complex high-frequency features while simultaneously enhancing the perceptual quality of the derained image. An effective Adaptive Fusion Module (AFM) is proposed for alternate feature aggregation. Extensive experiments have been done on both synthetic and real-world datasets, as well as extra raindrop benchmark RainDS and the downstream detection task. The experiment results have indicated that the proposed network exhibits more impressive performance and robustness than compared state-of-the-art methods.
\vfill
\pagebreak
\small
\ninept
\bibliographystyle{IEEEbib}
\bibliography{strings,refs}

\end{document}